\documentclass[sigconf, natbib=true]{acmart}
\usepackage{amsmath}
\usepackage{natbib}
\usepackage{amsmath}
\usepackage[toc,page]{appendix}
\usepackage{algorithm}
\usepackage{algpseudocode}
\usepackage{upgreek}
\usepackage{float}
\usepackage{multirow}
\usepackage{graphicx}
\usepackage[capitalize,noabbrev]{cleveref}

\usepackage{graphicx,subcaption}
\usepackage{enumitem}

\usepackage{balance}

\def\blfootnote{\gdef\@thefnmark{}\@footnotetext}

\newcommand{\youngsuk}[1]{\textcolor{blue}{[YS: #1]}}

\newcommand{\jonas}[1]{\textcolor{purple}{[JK: #1]}}

\newcommand{\yida}[1]{\textcolor{orange}{[Yida: #1]}}

\copyrightyear{2024}
\acmYear{2024}
\setcopyright{rightsretained}
\acmConference[KDD '24]{Proceedings of the 30th ACM SIGKDD Conference on Knowledge Discovery and Data Mining}{August 25--29, 2024}{Barcelona, Spain}
\acmBooktitle{Proceedings of the 30th ACM SIGKDD Conference on Knowledge Discovery and Data Mining (KDD '24), August 25--29, 2024, Barcelona, Spain}\acmDOI{10.1145/3637528.3671465}
\acmISBN{979-8-4007-0490-1/2}

\makeatletter
\gdef\@copyrightpermission{
  \begin{minipage}{0.3\columnwidth}
   \href{https://creativecommons.org/licenses/by/4.0/}{\includegraphics[width=0.90\textwidth]{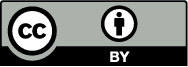}}
  \end{minipage}\hfill
  \begin{minipage}{0.7\columnwidth}
   \href{https://creativecommons.org/licenses/by/4.0/}{This work is licensed under a Creative Commons Attribution International 4.0 License.}
  \end{minipage}
  \vspace{5pt}
}
\makeatother

\begin{document}

\title{Inference Optimization of Foundation Models on AI Accelerators}


\author{Youngsuk Park
}
\affiliation{
AWS AI
  \city{Santa Clara}
  \country{USA}  
}
\authornote{Co-first authors. Correspondence: Youngsuk Park <pyoungsu@amazon.com>, Kailash Budhathoki <kaibud@amazon.com>. \\
	Tutorial website: \href{https://sites.google.com/view/kdd-2024-tutorial-inf-opt-/}{https://sites.google.com/view/kdd-2024-tutorial-inf-opt-/}
}
\author{Kailash Budhathoki\footnotemark[1]
}
\affiliation{
AWS AI
  \city{T\"ubingen}
  \country{Germany}  
}
\author{Liangfu Chen}
\affiliation{
AWS AI
  \city{Santa Clara}
  \country{USA}  
}
\author{Jonas K\"{u}bler}
\affiliation{
AWS AI
  \city{T\"ubingen}
  \country{Germany}  
}
\author{Jiaji Huang}
\affiliation{
AWS AI
  \city{Santa Clara}
  \country{USA}  
}
\author{Matth\"{a}us Kleindessner}
\affiliation{
AWS AI
  \city{T\"ubingen}
  \country{Germany}  
}
\author{Jun Huan}
\affiliation{
AWS AI
  \city{Santa Clara}
  \country{USA}  
}
\author{Volkan  Cevher}
\affiliation{
AWS AI
  \city{T\"ubingen}
  \country{Germany}  
}
\affiliation{
EPFL
  \city{Lausanne}
  \country{Switzerland}  
}
\author{Yida Wang}
\affiliation{
AWS AI
  \city{Santa Clara}
  \country{USA}  
}
\author{George Karypis}
\affiliation{
AWS AI
  \city{Santa Clara}
  \country{USA}  
}


\renewcommand{\shortauthors}{
Youngsuk Park \& Kailash Budhathoki et al.
}





\begin{abstract}

Powerful foundation models, including large language models (LLMs), with Transformer architectures have ushered in a new era of Generative AI across various industries. Industry and research community have witnessed a large number of new applications, based on those foundation models. Such applications include question and answer, customer services, image and video generation, and code completions, among others. However, as the number of model parameters reaches to hundreds of billions, their deployment incurs prohibitive inference costs and high latency in real-world scenarios. As a result, the demand for cost-effective and fast inference using AI accelerators is ever more higher. To this end, our tutorial offers a comprehensive discussion on complementary inference optimization techniques using AI accelerators. Beginning with an overview of basic Transformer architectures and deep learning system frameworks, we deep dive into system optimization techniques for fast and memory-efficient attention computations and discuss how they can be implemented efficiently on AI accelerators. Next, we describe architectural elements that are key for fast transformer inference. Finally, we  examine various model compression and fast decoding strategies in the same context.

\end{abstract}



\maketitle

\section{Overview}

The substantial size of modern large language models (LLMs), such as 
Llama-2/3 70B~\cite{touvron:2023:llama2}, Claude 3 Opus 137B~\cite{claude3}, and Groq-1 314B~\cite{grok1}, presents significant challenges in both training and inference phases. Training LLMs, in particular, demands considerable resources and has been the subject of extensive research. In contrast, inference consumes fewer computational resources but occurs much more frequently once the model has been trained. This phase is crucial as it encompasses various applications where the value of LLMs is realized, including text translation, sentiment detection, code generation, text summarization, and question answering.

Customers naturally demand faster and more cost-effective inference. To meet the user demands, it is essential to reduce 
latency---the time required to complete a generation---and to increase throughput, which is the number of requests processed per unit of time. The latency and throughput of LLMs depend on multiple factors, such as the hardware utilized, the capability of software frameworks to optimally leverage the available hardware, and the model architecture itself. Therefore, efforts to improve speed and costs benefit from optimizations across all these dimensions.
To this end, this section provides an overview of the characteristics of LLM inference, along with the corresponding systems and hardware requirements. 


\subsection{LLM Inference}
\begin{figure}[!htbp]
    \centering
    \includegraphics[width=.7\linewidth]{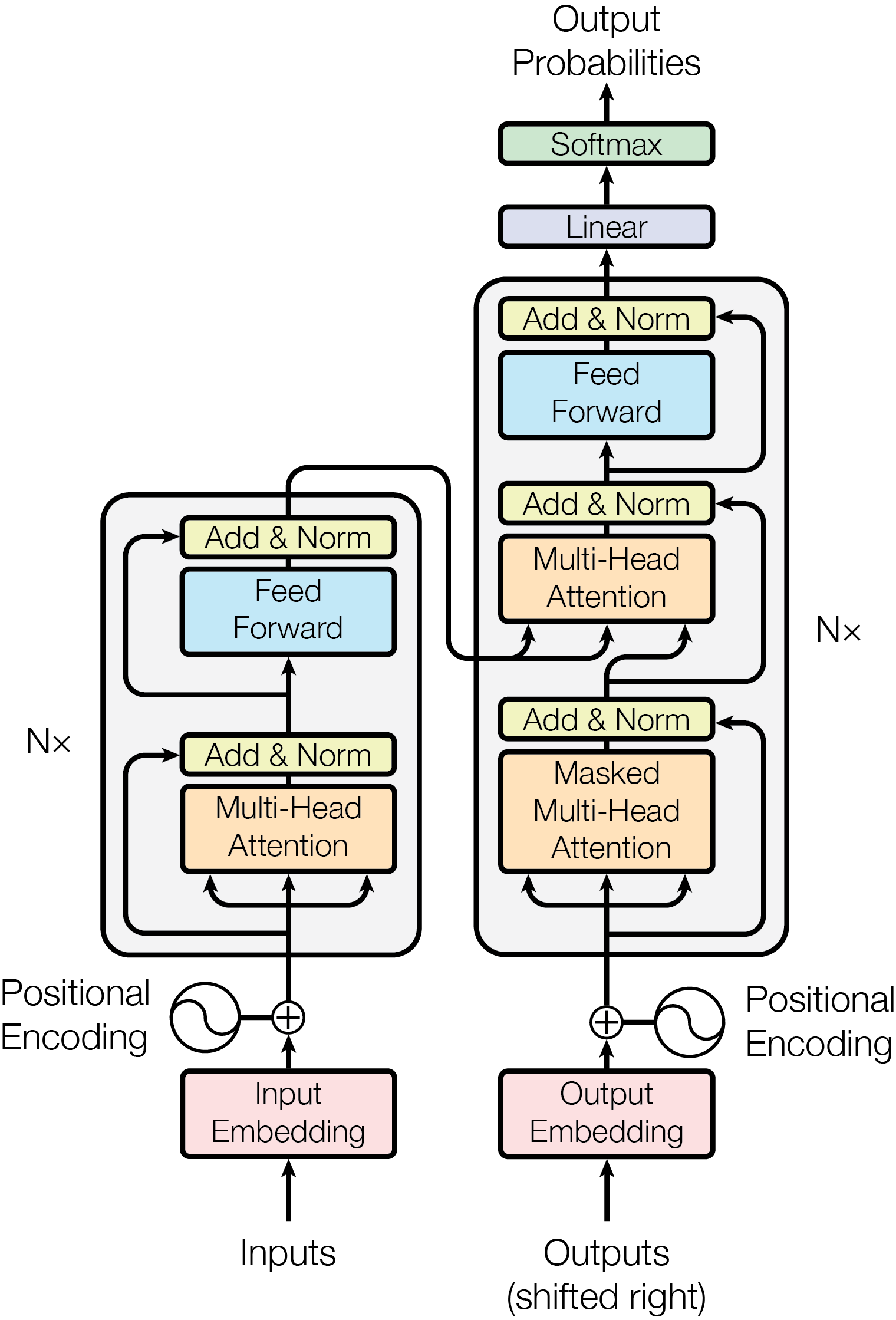}
    \caption{Original Transformer architecture adopted from \cite[Figure 1]{vaswani2017attention}, comprising of an encoder (left) and a decoder (right). Tokens are initially encoded into an embedding space and a positional encoding is used to encode information about the token positions. Modern LLM architectures are decoder-only with a backbone built of repeated layers containing masked attention and a feed forward neural network (FFN). The masked attention first applies linear transformations on a sequence of embeddings to obtain query ($Q$), key ($K$), and value ($V$) matrices and computes $\mathrm{Attention}(Q, K, V) = \mathrm{softmax}(\frac{QK^T}{\sqrt{d_k}})V$ thus relating the tokens to each other (the mask enforces that tokens can only attend to their predecessors). The FFN is applied on each token independently. Both attention and FFN add their outputs onto the embedding, which is passed through the skip connections. }
    \label{fig:transformer}
\end{figure}

Transformer models have revolutionized the landscape of LLMs by introducing a highly effective architecture for natural language processing tasks, as shown in \cref{fig:transformer} \cite{vaswani2017attention}. These models, characterized by their attention mechanisms, have significantly enhanced the capacity of models to understand and generate human-like text. Their versatility and scalability in training have established them as the backbone of many state-of-the-art LLMs today. Transformer models can include an encoding component only (e.g., BERT~\cite{devlin:2019:bert}), a decoding component only (e.g., GPT~\cite{brown2020language_gpt3}, Llama-2~\cite{touvron:2023:llama2}, Claude 3~\cite{claude3}, Groq-1~\cite{grok1}, Mistral 7B~\cite{jiang:2023:mistral}), or both (e.g., BART~\cite{lewis:2019:bart}). Currently, modern LLMs predominantly employ a decoder-only architecture, generating output sequences by predicting one token at a time, conditioned on the input sequence and previously generated tokens—a process known as auto-regression. Consequently, our discussion primarily focuses on decoder-only Transformer models.

As the model parameters of LLM increases, the decoding phase of LLM inference is inherently memory-bound due to its low arithmetic intensity, meaning that loading and moving the model weights into the on chip memory takes significantly more time than the actual computations.
This challenge becomes particularly acute with small batch sizes.
LLMs have a large memory footprint, primarily due to the pre-trained model weights and intermediate states required for next-token generation, such as the key-value cache. 

\subsection{Computational and Memory Requirements}
Modern computer chips employ specialized tensor units to efficiently perform tensor computations, such as matrix multiplication, which are fundamental in large foundation model workloads. Examples of these units include Nvidia TensorCore~\cite{tensorcore}, AMD MatrixCore~\cite{matrixcore},
and the systolic arrays found in Google  TPU~\cite{jouppi2017datacenter,jouppi2023tpu} and AWS  Trainium~\cite{bshara2024aws}. 
These tensor units are designed to process high-performance tensor computations such as matrix multiplication to meet the extensive demands of LLM workloads, especially during the training phase. 

Inference tasks, however, present a distinct challenge, as powerful tensor units alone are insufficient for optimal performance. 
To address memory-bound during decoding process, modern  chips incorporate high-bandwidth memory, typically in the form of Static Random Access Memory (SRAM). SRAM offers low latency and high throughput, suitable for the substantial memory requirements of inference workloads. However, the high cost of SRAM limits its capacity, requiring careful data manipulation to optimize its usage. 

\paragraph{High performance kernels} Inference-purposed kernels, such as DeepSpeed-Inference~\cite{aminabadi2022deepspeed}, FasterTransformer~\cite{fastertransformer_2022}, and transformers-neuronx~\cite{aws-neuron/transformers-neuronx_2024}, adhere to these guidelines to efficiently process the workloads. They can be designed by experienced performance-tuning experts or generated by machine learning compilers. In either case, a deep understanding of both chip architecture and inference workloads is essential for efficiently mapping and scheduling computations onto the hardware. By leveraging this knowledge, these kernels can fully optimize the utilization of high-bandwidth memory and tensor units, ultimately enhancing the efficiency of inference workloads on modern computer chips.

\paragraph{Hardware Accelerators}
While the majority of the LLM workloads are now done on GPUs following the SIMT (single instruction, multiple threads) paradigm,
LLM inference actually can also be accelerated with systolic array and High Bandwidth Memory (HBM) based systems (e.g. Google TPUs~\cite{jouppi2017datacenter,jouppi2023tpu}, AWS Trainium/Inferentia~\cite{bshara2024aws} and Intel Gaudi~\cite{gwennap2019habana}) with lower power consumption and lower cost accordingly.
Systolic array based systems can accelerate matrix multiplication with instruction-level parallelism ~\cite{jouppi2018domain}.
To accelerate memory access speed of a large amount of data, HBM is used as a replacement of Double Data Rate (DDR) and careful memory planning is required as the capacity of HBM is limited compared to the model size~\cite{zheng2020optimizing}.
There are also systems that utilize FPGAs~\cite{li2020ftrans} for compute acceleration, and systems that utilize inter-node connectivity \cite{zhong2024distserve} for large-scale transformer inference. 

\paragraph{Techniques to Mitigate Memory Bound}
In addition, to mitigate the memory-bound issues in LLM inference, practitioners employ various techniques that can be broadly categorized into two main approaches.
First, semantic-preserving methods aim to reduce memory usage while maintaining the original prediction via system optimization (\cref{sec:sys_opt}). Examples includes KV caches~\cite{pope2023efficiently}, FlashAttention~\cite{dao2022flashattention}, and FlashDecoding~\cite{flash-decoding_2023}. 
Conversely, architectural/algorithmic optimization usually trade off some prediction accuracy for improved memory efficiency and inference speed (\cref{sec:struct_transformer} and \cref{sec:model_compression}). These includes grouped query attention (GQA)~\cite{ainslie2023gqa}, Mixture of Experts (MoE)~\cite{shazeer2017outrageously} architectures as well as compression methods of quantization, pruning and distillation, and speculative decoding~\cite{chen:2023:speculative}. 

\subsection{Distributed Solution Frameworks}
The memory-bound nature of LLM inference and the limited capacity of HBM on individual accelerators present significant challenges in meeting the growing demands of LLM workloads. LLMs with hundreds of billions of parameters typically do not fit on a single node for inference, let alone a single accelerator. Consequently, a distributed solution becomes necessary. However, implementing such a solution for LLM inference introduces challenges like efficient model partitioning, communication, and load balancing. Addressing these challenges is crucial for enabling scalable processing of large-scale LLM inference workloads. 
Typically, we can employ a combination of multiple parallel strategies to achieve state-of-the-art performance for LLM inference, each with its own advantages and disadvantages. 


\textbf{Tensor parallelism} is designed to distribute large chunks of tensor computation workloads across multiple accelerators and aggregate the final results via collective communication. This approach can help reduce end-to-end latency when collective communication is efficient (e.g., NVIDIA NVLink \cite{foley2017ultra}, AWS Neuron Collective Communication \cite{aws_2024}). However, if the tensor computation workload is small, the extra overhead in collective communication can diminish overall performance. Since inter-node communication is typically higher than intra-node communication, tensor parallelism is most effectively utilized within a single node.

\textbf{Pipeline parallelism} is employed to distribute model layers across accelerators. As both model weights and KV cache for each layer can be distributed to different accelerators, and only the inputs/outputs of the layers need to be transferred across devices, pipeline parallelism is relatively independent of the collective communication bandwidth. This strategy allows for the distribution of models that are too large for a single node. To increase hardware utilization, overlapping different pipeline stages is typically necessary. Pipeline parallelism is preferable over tensor parallelism when the entire model does not fit on a single node for inference.


\textbf{Sequence parallelism}~\citep{l:2022:sequence-parallel} is a critical technique for supporting long context. The core concept of sequence parallelism involves distributing sequences along the sequence dimension, enabling the parallel decoding of small batches of long sequences. This technique is implemented by solutions such as FlashDecoding \cite{flash-decoding_2023}
and PagedAttention V2 \cite{kwon_2023}.

\textbf{Expert parallelism} (EP) facilitates the distribution of Mixture of Expert (MoE) models~\cite{shazeer2017outrageously}
across multiple accelerators. The MoE model architecture is designed to skip inactive expert computation, while still maintaining the capability to achieve high accuracy compared to dense models. Since expert weights are typically large, distributing and dynamically loading these weights can be costly. To reduce collective communication and avoid the dynamic loading of expert weights, EP keeps each expert within a small group of accelerators \cite{rajbhandari2022deepspeed}. As the input/output data is considerably smaller than expert weights, the all-to-all collective communication can be efficiently used to distribute tokens to the activated experts.

\section{System Optimization }\label{sec:sys_opt}
We explores semantic-preserving optimizations for LLM inference from a systems perspective. By strategically organizing computations, significant improvements in inference speed and memory efficiency can be achieved without compromising the semantic integrity of the model. In this seciton, we discuss on reducing redundant computations through the use of key-value caches (\cref{sec:KV-Cache}), optimizing attention implementation to minimize memory access (\cref{sec:flash-attention}), enhancing throughput via handling batches of requests (\cref{sec:continuous_batching}), and reducing unused memory fragmentation via distributing sequences (\cref{sec:paged_attention}). 
These optimizations were mainly developed based on GPUs, but the main concepts are largely applicable to other AI accelerators with some specific implementation tweak.
The following subsections delve into each of these approaches in detail, examining their theoretical foundations, practical implementations and challenges therein.

\subsection{Fast Attention Computation via Caching}\label{sec:KV-Cache}

Generating tokens in an autoregressive fashion is a widely adopted approach like GPT~\cite{brown2020language_gpt3} and Llama~\cite{touvron:2023:llama2}, yet it can pose computational challenges.
During the auto-regressive generation, decoding step to generate every next token requires to fetch  previous tokens. This requires to compute their hidden representation of keys and values in attention mechanism, which could be repetitive during the sequence of token generation. KV- cache~\cite{pope2023efficiently} stores and reuses these past key-value pairs, eliminating the need of recalculation for every new token.
This technique significantly improves the efficiency of inference, by reducing the quadratic complexity of attention computation w.r.t. a sequence length to be linear. 


However, the memory footprint of KV cache growing linearly w.r.t. the sequence length can be substantial, as it requires additional memory to store the cached keys and values. To address this, several techniques has been introduced to reduce the memory space required for the KV cache. Low-bit precision data types have been utilized in KVQuant~\cite{hooper2024kvquant}, which brings million-scale context length support on a single A100-80G GPU. StreamingLLM~\cite{xiao2023efficient} introduced the concept of attention sink, which preserves decent accuracy by leveraging initial tokens without exhausting the long context window size. 
Generalized block-sparse attention patterns, e.g. BigBird~\cite{zaheer2020big}), allow the training of long context support, without degrading accuracy at inference stage. Heavy-Hitter Oracle~\citep{zhang:2023:h2o} is a cache eviction policy which retains Heavy Hitters tokens, i.e., tokens contributing most of the value in attention scores, based on local statistics at each decoding step. However, all of these can lead to a potential degradation of accuracy. 

The aforementioned KV cache strategies can be implemented differently depending on hardware. To be specific, 
the KV cache memory space size can be formulated as $2bshdln$ bytes, 
where $b$ is batch size, $s$ is sequence length, $h$ is number of KV heads, $d$ is size of the attention head, $l$ is the number of layers, $n$ is size of each data element in number of bytes.
The size of $b$ is determined at runtime for batch inference.
$d$ and $l$ are fixed by the model configuration.
This leaves the optimization space for reducing KV cache memory space being limited to $s$, $h$ and $n$.
KV cache quantization helps the reduction of $n$. Block-sparse attention techniques help minimize $s$ and $h$. With all considered, the distributed strategy of KV cache memory can be distinct among GPU and systolic array-based accelerators (e.g., TPU, Trainium) due to different memory constraints and the numbers of devices per node, especially for handling GQA models (Section \ref{sec:MQA_GQA}).

PagedAttention~\cite{kwon2023efficient} can be considered as a KV cache optimization. It transforms the KV cache into non-contiguous memory space, and makes $s$ as a fixed block size. 
Each sequence can occupy a variable number of KV cache blocks. SGLang~\cite{zheng2023efficiently} further transforms the fixed block size to variable length, with RadixAttention enabling automatic KV cache reuse.
Both PagedAttention and RadixAttention enabled the possibility to cache shared prefix among multiple sequences, without duplicated copy of the prefix.






\subsection{Efficient Attention Computation}\label{sec:flash-attention}

\begin{figure}[!htbp]
    \centering
    \includegraphics[width=0.9\linewidth]{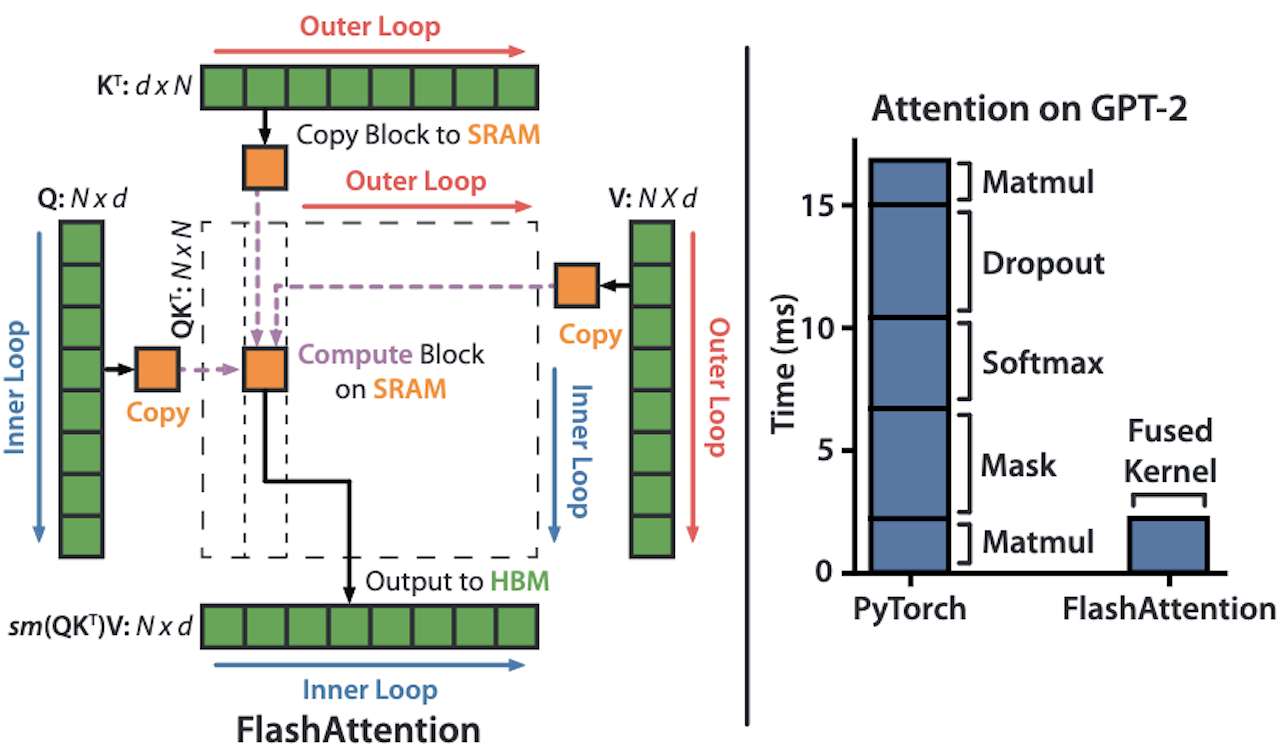}
    \caption{Flash Attention by \citet{dao2022flashattention}. The outer loop iterates over K and V blocks and loads them to fast SRAM. In each block, inner 
loops iterates over Q blocks, loading them to SRAM, and writing the attention output back to HBM.
}
    \label{fig:enter-label}
\end{figure}
Modern LLMs have extended the support of context length from the order of thousands to millions within a few years from less than 1k (e.g., GPT-2~\cite{brown2020language_gpt3}) to 200k+ (e.g., Claude 3 \cite{claude3}). 
The main challenge of expanding the context window lies in the extensive computational requirements and memory consumption for the attention computation. As the model considers more tokens simultaneously, the compute/time complexity and memory demands of calculations increase significantly, scaling quadratically with the size of the context window. 
FlashAttention~\cite{dao2022flashattention, dao2023flashattention} was introduced to address these challenges, which reformulates the attention computation as a sequence of matrix multiplications and applies block-sparse decomposition. By processing attention in smaller blocks, FlashAttention reduces the memory footprint of attention computation, avoiding the need to materialize the entire attention matrix in memory at once.
The key advantage of FlashAttention is its ability to minimize data movement between different memory hierarchies. 
By carefully selecting the block size based on the memory hierarchy and capacity of the device, FlashAttention ensures that the data can be efficiently processed without requiring multiple transfers between memory levels. For example, on GPUs, the block size is typically small to fit within the L2 cache, minimizing expensive memory accesses. In contrast, devices like AWS Trainium or Google TPU, which have a large scratchpad memory in the tens of megabytes (MBs), can leverage larger block sizes to maximize computational efficiency by processing more data in parallel.

For large context, Blockwise Parallel Transformer (BPT) \cite{liu2024blockwise} further minimize memory consumption on feedforward network by computing them in a block-wise manner. Enhancing BPT, Ring Attention \cite{liu2023ring} utilizes blockwise computation for self-attention and feedforward processes to distribute extended sequences across multiple devices by dividing the input text into smaller, more manageable blocks. These blocks are processed on separate devices organized in a ring-like configuration, enabling parallel processing. 

When it comes to inference compared with training, relatively smaller batch size can lead to different bottleneck. Flash-Decoding \cite{flash-decoding_2023}, based on FlashAttention, introduces a new parallelization dimension: the keys/values sequence length. 
It stores minimal extra data in global memory while fully utilizing the accelerator, even with small batch sizes, provided the context length is sufficiently large. For the smaller chunks of split keys/values, it computes the attention of the query with each chunk in parallel using FlashAttention, and reduce across all chunks to calculate the final output.

\subsection{Continuous Batching}\label{sec:continuous_batching}

\begin{figure*}[!htbp]
    \centering
    \includegraphics[width=\textwidth]{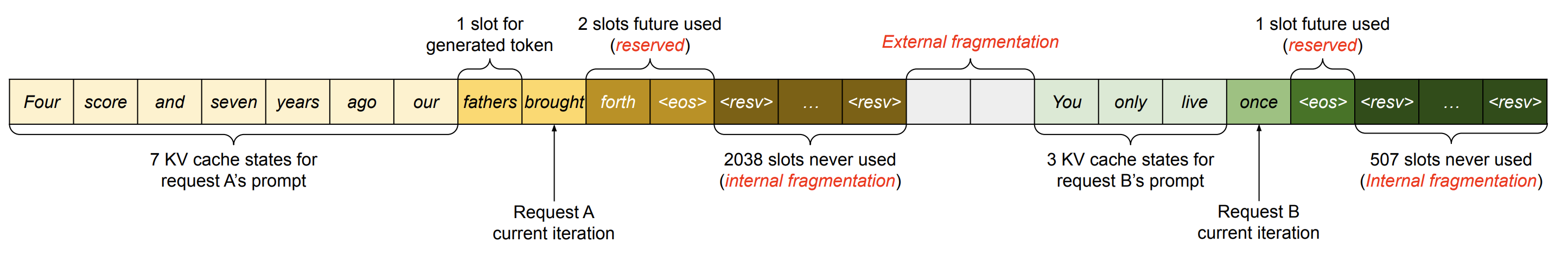}
    \caption{Types of memory fragmentation by \citet{kwon2023efficient}. The figure depicts the memory space for decoding two sequences.
    Internal memory fragmentation is considered to be the allocated KV cache blocks that are not occupied by the sequences.
    The free memory space that is not allocated is considered to be external memory fragmentation.
    }
    \label{fig:types-of-memory-fragmentation}
\end{figure*}

LLM inference is inherently memory-bound if only one sequence is processed.
To increase the throughput for a large number of input prompts, the most straightforward approach was to 
allocate a fixed time window for decoding a fixed number of sequences.
This is commonly known as static batching, which has been implemented in FasterTransformer \cite{fastertransformer_2022} and many others \cite{pope2023efficiently, aws-neuron/transformers-neuronx_2024}.
The advantage of static batching comes from the minimized latency for decoding with small batch sizes.
As batch size gets bigger to achieve higher throughput, a mechanism in improving effective utilization of batched decoding is needed.

Static batching results in resource waste as some sequences reach the end earlier than the others in the same batch.
Orca \cite{fossati2022distributed} proposed the idea of a dynamic sequence eviction strategy.
The strategy essentially removes the sequences that generated EOS token, and inserts new prompts into the decoding batch.
The approach is commonly referred to as continuous batching. 
In addition to the proposed mechanism in handling continuous batching, Ocra also introduced the idea of flattening multiple input prompts
and concatenate them into the prefill kernel, in order to reduce padding and kernel launch overhead.
The block diagonal causal attention mask is commonly used to achieve a throughput gain with FlashAttention.

\subsection{PagedAttention and its Derived Applications}\label{sec:paged_attention}

Since the length of output tokens is unpredictable, the most straightforward approach was to maintain the maximal sequence length for each decoding request.
As most part of the reserved memory won’t be actually used, this would introduce a large amount of internal memory fragmentation. 
As illustrated in the Figure \ref{fig:types-of-memory-fragmentation}, internal memory fragmentation refers to the memory space that is allocated but not effectively utilized for sequence decoding.
External memory fragmentation indicates the device memory space that is free but not allocated for usage.
To reduce both internal and external memory fragmentation, PagedAttention \cite{kwon2023efficient} introduced the idea of distributing sequences in non-contiguous physical memory space.
The capability of serving LLM with PagedAttention has been demonstrated with the vLLM project \cite{vllm-project/vllm_2024}, NVIDIA TensorRT-LLM \cite{nvidia/tensorrt-llm_2024}, and HuggingFace TGI \cite{huggingface_2023}.
Since the initial software release of the vLLM project, a few extensions have been added as part of the improvement:

\begin{itemize}[leftmargin=*]
    \item \textbf{FP8 (E5M2/E4M3) data type \cite{micikevicius2022fp8} for KV cache storage}. ~
    FP8 storage data type for the KV cache helps increase compute intensity in decoding stage, and mitigate the memory-bound decoding problem.
    It can also help increase batch size while maintaining same amount of KV cache payload, comparing to FP16/BF16 KV cache data type.
    The throughput benefit can come from the increase of batch size for decoding.
    The initial support for FP8 KV cache quantization~ \cite{yang_2023} in vLLM reported 1.49x throughput improvement on A100, by trading off up to 2.4\% of accuracy degradation on HumanEval-Python evaluation tasks.
    \item \textbf{Structured KV cache storage for shared prefix processing}.~
    Recent advancement in context-aware generation has demonstrated strong reasoning capability in multiple frameworks \cite{Chase_LangChain_2022, guidance-ai_2023, khattab2023dspy}.
    To reduce unnecessary computation while maintaining strong reasoning capability, 
    \citet{zheng2023efficiently} 
    proposed RadixAttention, which utilize radix tree and maintain the tree elements as sequences with varying lengths.
    It also introduces a compiler optimization framework to achieve longer shareable prefixes for caching.
    \item \textbf{Reduce the interruption of input prompt encoding.}~
    In order to reduce high tail latency in decoding phase due to long context inputs,
    \citet{agrawal2024taming} proposed the idea of distributing long context inputs into separate chunks of processing steps.
    It utilizes the chunked prefill kernel, which was initially proposed to reduce pipeline bubble for multi-GPU serving.
    It increases the stability in decoding latency via stall-free decoding, and improved end-to-end throughput by up to 1.33x.
\end{itemize}






\section{Structured Transformer Architectures}\label{sec:struct_transformer}

Beyond optimizing the serving of a given model, also the model architectures themselves have developed and moved towards architectures that enable faster and more efficient inference, while still being similarly powerful. In the following we discuss changes to the attention mechanism, reducing its number of key and value heads (\cref{sec:MQA_GQA}) as well as mixture of experts approaches, which effectively only execute part of the network for each token (\cref{sec:moe-section}), in addition to other architecture choices (\cref{sec:other_arch}).


\subsection{Multi-/Grouped Query Attention} \label{sec:MQA_GQA}

Falcon~\cite{almazrouei2023falcon} and Llama 2 70B~\cite{touvron:2023:llama2} employ techniques known as multi-query attention (MQA) \cite{shazeer2019fast}  and grouped-query attention (GQA) \cite{ainslie2023gqa} respectively. When it comes to inference, memory and computational challenges arise from the repeated loading of decoder weights and attention keys/values in decoding steps. 

\begin{figure}[!htbp]
    \centering
    \includegraphics[width=0.9\linewidth]{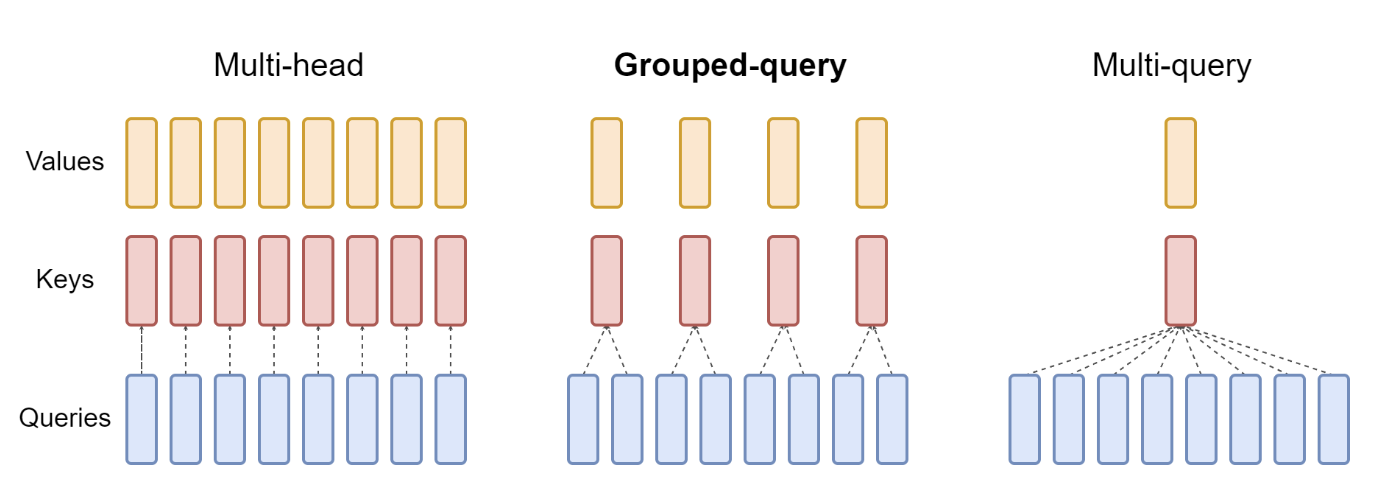}
    \caption{Overview of grouped-query method by \citet{ainslie2023gqa}. }
    \label{fig:GQA}
\end{figure}

In multi-head attention, distinct queries brings linear increase on the number of heads for keys and values, requiring larger memory bound and prohibiting potential latency improvement. However, MQA involves employing multiple query heads alongside a single key/value head, thereby accelerating decoder inference. GQA an advancement over MQA, strikes a balance by utilizing an intermediate number of key-value heads (more than one but fewer than the query heads). The GQA model efficiently partitions the query into $n_\text{heads}$ segments akin to the original multi-head attention mechanism, while dividing the key and value into handful of groups. For example, Llama-3 70B \cite{touvron:2023:llama2} uses 64 query heads which are grouped onto 8 key-value heads. This arrangement allows a handful of query heads to share the same key-value heads to interact. 
By leveraging repeated key-value pairs, the GQA approach enhances overall model performance while preserving quality.

When it comes to the MQA/GQA inference strategy in a distributed setting, there are a number of approaches.
If possible, the common practice is to evenly distribute KV heads across multiple accelerators.
This assumes that the number of KV heads are divisible by the number of accelerators.
For handling the case where there are more accelerators than number of KV heads,
\citet{pope2023efficiently} has introduced the approach that distributes sequences over different accelerators. 
The idea is to leverage all-to-all operator to transform the layout of the hidden states.
It can effectively increase static batching inference throughput when batch size is large and number of KV heads is small.
Regarding the support of PagedAttention \cite{kwon2023efficient}, the best practice is yet to be explored, since KV cache block placement is determined at runtime. It won't be effective if number of KV cache blocks are imbalanced among accelerators. Existing solutions either shard along sequence dimension (e.g. PagedAttention V2), or replicate the KV heads for each of accelerators.


\subsection{Mixture of Experts for Transformer}
\label{sec:moe-section}

\begin{figure}
    \centering
    \includegraphics[width=\linewidth]{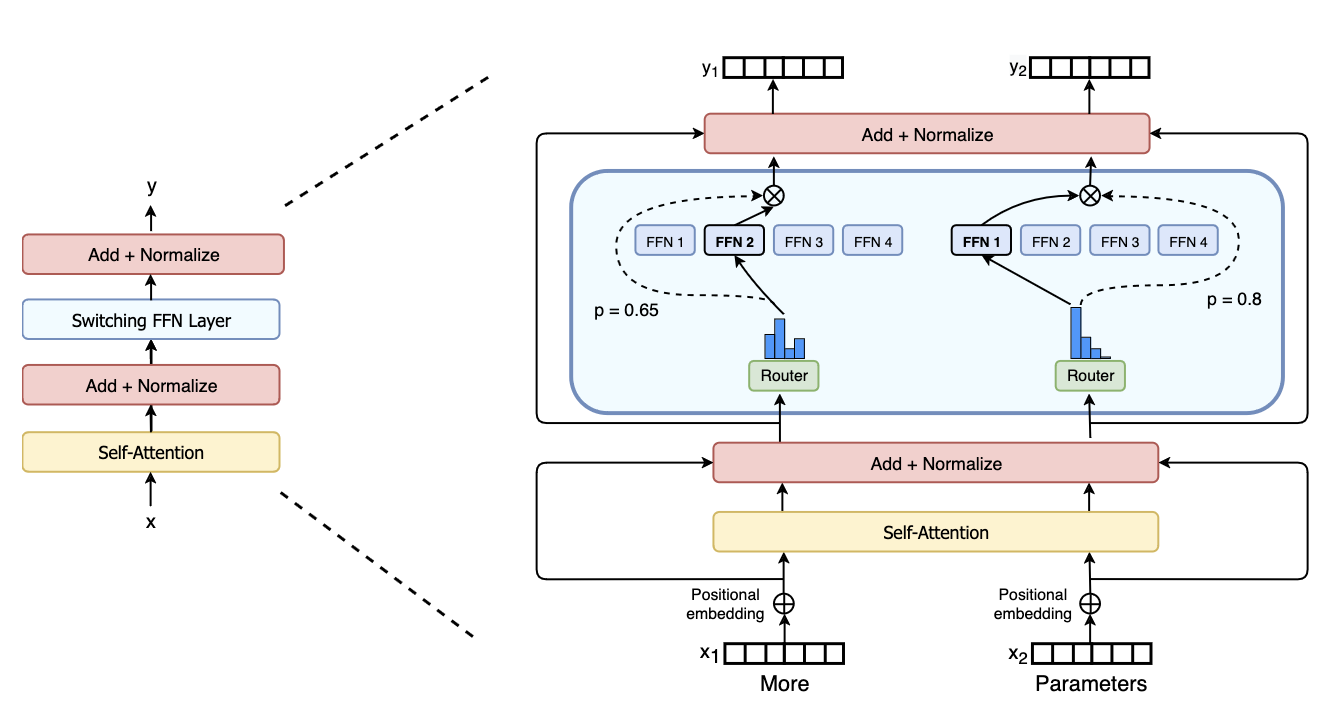}
    \caption{Instead of the dense feed-forward network layer in the traditional Transformer (left blue), \citet{fedus2022switch} introduce a sparse Switch FFN layer (right blue). This layer functions independently on the sequence's tokens.}
    \label{fig:moe}
\end{figure}

Mixture of Experts (MoE)~\cite{shazeer2017outrageously} architecture from Figure~\ref{fig:moe} is designed to activate part of expert computation by skipping inactive ones, while maintaining the capability in achieving high accuracy. This allows for pretrained models to utilize significantly less computational resources and thus increase the model's size or the dataset it handles within the same computational budget compared to a dense model both in training and inference. This MoE component becames a popular design choice in favor of fast inference among Transformer class 
\cite{lepikhin2021gshard, fedus2022switch, jiang:2023:mistral}. Among many variance of MoE \cite{zhang2021moefication, gao2022parameter, komatsuzaki2022sparse, lewis2021base, zhou2022mixture, zoph2022st}, it typically tries to comprise two primary components:
First, sparse MoE layers replace conventional dense feed-forward network (FFN) layers. These MoE layers are comprised of a set number of "experts" (e.g., 8 in Mistral~\cite{jiang:2023:mistral}), where each expert functions as an individual neural network. While these experts are typically FFNs in practice, they can also encompass more intricate networks or even form a hierarchical MoE structure~\cite{fritsch1996adaptively}.
Second, a gate network or router determines the allocation of tokens to specific experts. Notably, tokens can be directed to multiple experts. This decision is governed by the routing mechanism as a critical design choice for efficient inference and training. The router, comprising learned parameters, is pretrained concurrently with the remainder of the network and plays a pivotal role in token allocation within MoEs. 

Routers for sparse MoEs can be categorized into two main variants: Token Choice, which assigns experts to individual tokens, and Expert Choice, which assigns tokens to individual experts. 
Token Choice can be optimal for latency constrained applications, since the number of activated experts is small.
Expert Choice is used for throughput optimizations, especially when total number of experts are small and the tokens can be balance among all experts.
In such applications, Expert parallelism (EP) keeps each expert within a small group of accelerators, leading fast inference by alleviating collective communications and dynamic loading.



\subsection{Other Architectures}\label{sec:other_arch}
Sliding Window Transformer (SWT)~\cite{beltagy2020longformer} is a variant of the self-attention mechanism designed to handle long sequences more efficiently by dividing the input sequence into smaller, overlapping chunks or "windows." For each token, the attention score is computed only over a window of length $w$ sequence rather than the entire (previous) sequence. This attention mechanism sequentially slides across the input sequence to compute all localized attention scores. As the layers of the SWT get deeper, the localized attention mechanism extends the receptive fields w.r.t. input tokens, preserving a comprehensive understanding of the entire sequence, similar to a CNN. Each SWT requires only linear complexity $O(nw)$, mitigating the quadratic complexity $O(n^2)$ in standard self-attention.


Mixture-of-Depth~\cite{raposo:2024:mixtureofdepths} allows some tokens to take paths across layers dynamically, skipping certain layers based on specific criteria, e.g., CALM~\cite{schuster:2022:calm} with exit criteria during forward pass, instead of all tokens passing through every layer of Transformer. This approach enables the model to allocate computational resources more efficiently, focusing more layers on complex parts of the input while using fewer layers for simpler parts. The mixture of depth can help reduce computational costs and improve forward/backward speed without significantly compromising model performance. 

\section{Model Compression}\label{sec:model_compression}
Model compression techniques~\cite{chittyvenkata:2023:inf-survey} compress a model or input, thereby reducing the memory footprint and latency of LLMs. These methods come with challenges as they typically introduce trade-offs between inference improvement and accuracy.
Quantization of  model weights (\cref{sec:quantization}) has essentially has become a standard nowadays. Pruning parts of models has posed more challenges but also seen much progress targeted specifically to LLMs (\cref{sec:pruning}). Lastly, entirely compressed models can be trained through distillation from a large teacher model (\cref{sec:distillation}).

\subsection{Quantization}\label{sec:quantization}
\begin{figure}[!htb]
    \centering
    \includegraphics[width=0.75\columnwidth]{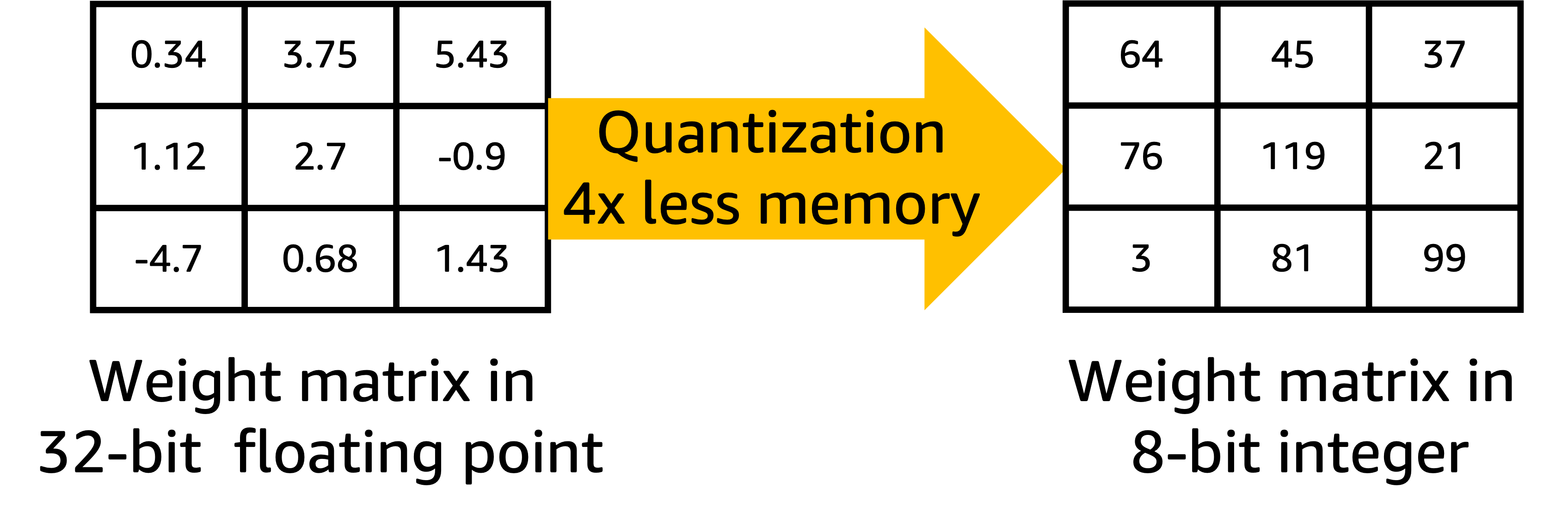}
    \caption{INT8 quantization~\cite{xiao:2023:smoothquant,dettmers:2022:llmint8} represents weights or activations in FP32 data types into 8-bit integers.}
    \label{fig:quantization}
\end{figure}

Quantization~\cite{gholami:2021:quantization-survey} is a model-compression technique that represents weights or activations of the network with low-precision data types (e.g., 8-bit integer) instead of high-precision data types (e.g., FP32), therewith reducing the storage when loading the model/activations in hardware (see Figure~\ref{fig:quantization}). Reduced data precision poses trade-off between latency-throughput-accuracy. It also requires support from the target hardware to realize maximum speedup~\citep{vanbaalen:2023:fp8vsint8}. Quantization is applied either during training or after training.

\emph{Post-training quantization (PTQ)} methods quantize  weights or activations of a pre-trained model (e.g., LLM.int8() \citep{dettmers:2022:llmint8}, ZeroQuant-V2~\citep{yao:2023:zeroquantv2}, SmoothQuant~\cite{xiao:2023:smoothquant}, GPTQ~\citep{frantar:2023:gptq}, Quip\#~\citep{tseng:2024:quip}, OmniQuant~\citep{shao:2024:omniquant}, AQLM~\citep{egiazarian:2024:aq}, PV-Tuning~\citep{malinovskii:2024:pvtuning}, Outlier Suppression+~\citep{wei:2023:outlier}, QLoRA~\citep{dettmers:2023:qlora}). 
Additional fine-tuning is often needed to recover the downstream accuracy drop~\citep{kwon:2022:alphatuning}.

\emph{Quantization aware training (QAT)} emulates inference-time quantization, creating a model that can be quantized later post-training (e.g., 1-bit LLM~\citep{wang:2023:bitnet}, 1.58-bit LLM~\citep{jacob:2017:quantization, ma:2024:1-bit-era}, LLM-QAT~\citep{liu:2023:llmqat}, QLLM~\citep{liu:2024:qllm}). They stipulate full pre-training of the base model. A promising recent work~\citep{ma:2024:1-bit-era} reported accuracy of a model with only ternary $\{-1,0,1\}$ weights on par with a full-precision model.

The upper bound for both latency and throughput improvement from weight-only quantization is the ratio of source precision data type to the target precision data type. For example, the upper bound for latency/throughput improvement with INT8 quantization down from 32-bit floating point format (FP32) is $4\times$. As INT8 parameters require $4\times$ fewer bits than FP32, we can increase the batch size as well as perform more computations on the same data size in one go. But memory saving does not directly translate to improved throughput/latency due to several factors like memory bandwidth, hardware limitations, and quantization/de-quantization overhead.




\subsection{Pruning}\label{sec:pruning}
Pruning is a compression technique to remove redundant parameters from the model. The goal is to maintain prediction quality of the model while shrinking its size, and thereby increasing its efficiency. Pruning requires strategies to identify which parts to remove and, potentially, how to adapt the remaining parts in order to compensate for quality degradation.

\emph{Structured Pruning} removes whole components of the network such as neurons, attention heads, and layers \cite{ma2023llmpruner, kurtic2023zipLM, ashkboos2024slicegpt, xia2022Structured,tao-etal-2023-structured,loraprune_Zhang2023}. For example, \emph{SliceGPT} \cite{ashkboos2024slicegpt} effectively decreases the embedding dimension of the model, whereas \emph{LLM-Pruner}~\cite{ma2023llmpruner} scores coupled structures in the decoder-layer and removes the 
least important 
ones.  
\emph{Sheared Llama}~\cite{xia2024sheared} and \emph{LoRAPrune}~\cite{loraprune_Zhang2023} similarly remove entire structures. When pruning larger structures like channels, blocks, or embedding dimensions, the speedups can easily be realized end-to-end 
(e.g., \cite[Table~3]{ma2023llmpruner},  \cite[Table~2]{ashkboos2024slicegpt}).

\emph{Unstructured Pruning} removes individual weights of the network. Clearly, weights that are 0 can be ignored without any loss in accuracy, but also very small weights can be set to zero. Pruning weights that are not small enough will finally lead to degradation of the model, which sets the limit for the speedup. Given a desired sparsity ratio and a matrix~$W$, the simplest strategy is to prune the weights with the smallest magnitude, which corresponds to minimizing the Frobenius norm between the dense matrix~$W$ and its sparse approximation $\widehat{W}$, i.e.,  $\|W-\widehat{W}\|_F^2$. 
This approach, referred to as \emph{magnitude pruning}, 
quickly leads 
to drastic accuracy degradation \cite{frantar2023sparsegpt, sun2023wanda}.   \emph{Wanda} \cite{sun2023wanda} and \emph{RIA} \cite{pap_Zhang2024} improve over  
simple magnitude pruning 
by reweighing the matrix weights with the norm of the corresponding input activation. Another popular Transformer pruning method is \emph{SparseGPT}\cite{frantar2023sparsegpt}, which jointly optimizes the pruning mask as well as the remaining weights in order to minimize $\|(W-\widehat{W})X\|_F^2$, where $X$ represents a sample of inputs to the linear layer. Since finding the optimal pruning mask is a combinatorial problem, SparseGPT employs heuristics to make it computationally feasible. While most methods apply the sparsity uniformly across layers, \emph{owl} \cite{yin2023outlier}, BESA \citep{BESA_Xu2024}, and \emph{ISC} \citep{ISC_Shao2023} derive a criteria to prune layers to different levels.

Unstructured sparsity is mainly of academic interest since, so far, it does not lead to speedup on hardware accelerators (\emph{Flash-LLM} \cite{xia2023flash} recently provided some steps in this direction). However, 
most 
methods can also be applied to achieve \emph{N:M structured sparsity}, 
where only $N$ out of $M$ consecutive elements are allowed to be non-zero.  
Some hardware accelerators support these patterns and allow for memory savings and speedups~\cite{nvidia2021structured}. 

While pruning can in principle be done during pretraining \cite{lasby2024dynamic, zhu2017prune,platon_Zhang2022}, most recent work focuses on the post-training setting. Nonetheless, in order to recover from the accuracy loss due to pruning many works consider applying a short training strategy after pruning. This is either done via a standard pretraining loss \cite{sun2023wanda, ma2023llmpruner} or with variants of distillation losses \cite{sanh2020movement, lagunas2021block, kurtic2023zipLM, xia2024sheared,tao-etal-2023-structured}. To increase efficiency, some works do not update all remaining parameters, but employ parameter efficient techniques like LoRA \cite{hu2021lora}. Generally, such strategies help recovering the accuracy loss, but are also prone to overfitting to the specific dataset used \cite{ma2023llmpruner} and can compromise the generality of the model.


\subsection{Distillation}\label{sec:distillation}
\begin{figure}[!htb]
    \centering
    \includegraphics[width=0.5\columnwidth]{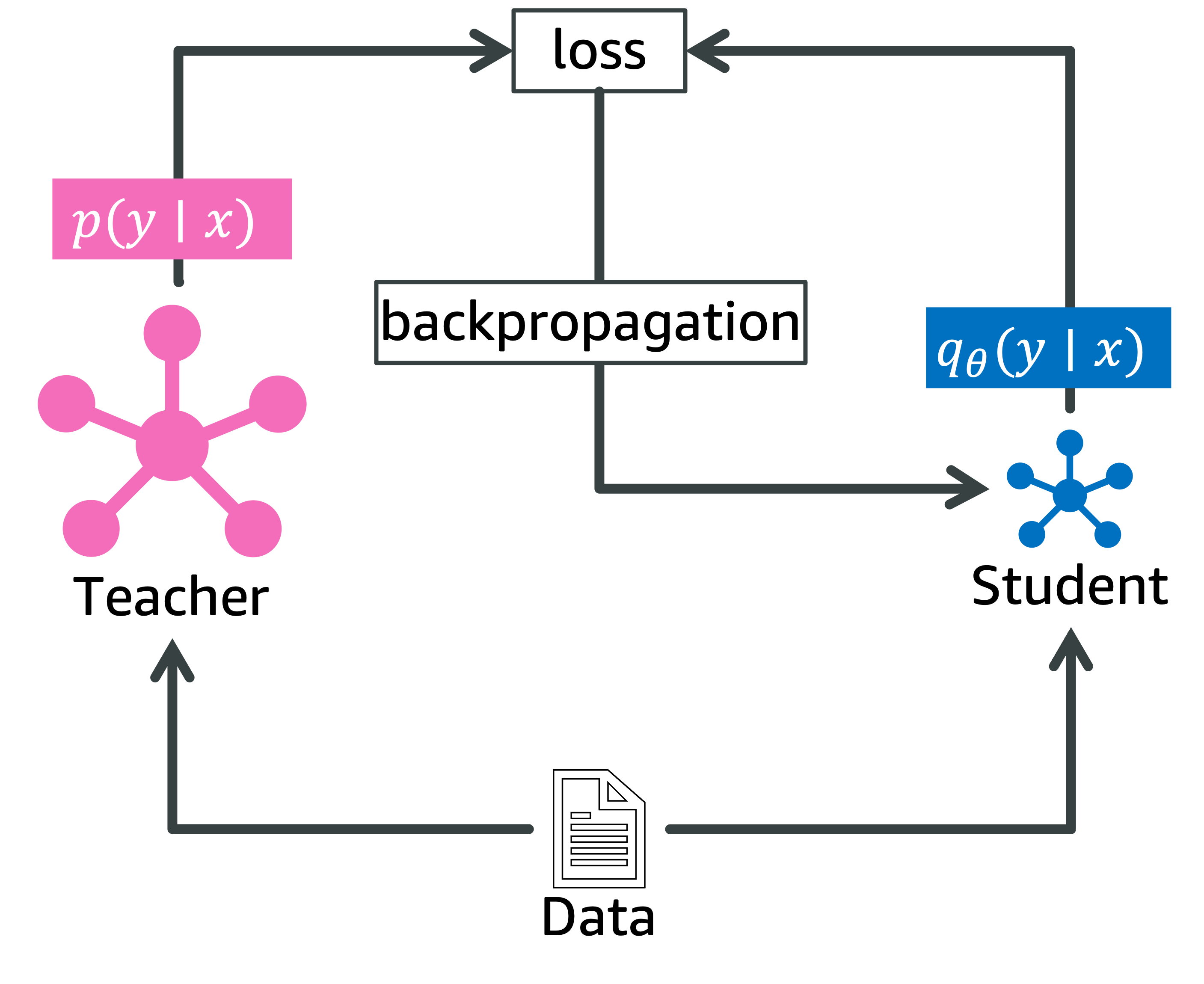}
    \caption{Canonical knowledge distillation process by \citet{hinton:2015:distilling}, where small student model distills a large teacher model via minimizing a distillation loss. This loss on a transfer dataset is then backpropagated to the student model.}
    \label{fig:distillation}
\end{figure}

Knowledge distillation (KD) (\cite{bucila:2006:model-compression, hinton:2015:distilling,west:2022:symbolic,lopes:2017:datafree}) is a model compression technique in which we train a small model (called student) to match closely the performance of a larger model or an ensemble of models (called teacher). To this end, KD connects a student model with the teacher model by a distillation loss, which penalizes differences in the outputs of the two models at certain layers (see Figure~\ref{fig:distillation}). The standard KD approach—also called last-layer-only approach—trains the student to match the performance of the teacher at the last layer (e.g., ~\cite{hinton:2015:distilling, sanh:2020:distilbert}). Another 
approach—also called layer-wise approach—trains the student to match the hidden representation of the teacher at each layer (e.g.,~\cite{sun:2019:bert-distill-layerwise}). Layer-wise-distillation approaches report improved results on downstream tasks compared to last-layer-distillation approaches~\citep{liang:2023:less-is-more}, but they stipulate the same number of layers in the student as the teacher. In general, KD approaches are flexible with regard to the exact structure of the student model, which allows optimizing the student for various target hardwares. Another advantage is that the distillation process runs entirely after training the large teacher model.

Distillation does not affect the training of a teacher model, but distillation effort by itself can be a major training effort for the following reasons. First, the number of steps can be similar to pre-training a small model. Second, the distillation loss usually is a combination of the pure student/teacher loss together with an original loss, for which typically the original pre-training data is recommended~\cite{hinton:2015:distilling}. To compute the distillation loss, we also need to make a forward pass of a teacher model to get logits.
But there is a range of possibilities in selecting the transfer set on which to train the smaller distilled model~\citep{peris:2022:knowledge}. For example, symbolic distillation~\citep{west:2022:symbolic, liu:2023:llmqat} approaches synthesize data from the teacher model to this end. Distillation also comes with a trade-off between size and quality, which determines the improvement in throughput/latency.




\section{Fast Decoding}

As discussed, vanilla auto-regressive decoding is memory bound. Speculative decoding (SD)~\cite{Leviathan2023,chen:2023:speculative} exploits the fact that multiple draft tokens can be verified in a single forward pass of the target model. The draft tokens are then accepted based on a rejection sampling scheme~\cite{Leviathan2023,chen:2023:speculative} or deterministic approaches~\citep{Kim2024}. Processing the draft tokens requires additional computations in the target model, but the main bottleneck remains the loading of the weights. Hence, the verification of the additional draft tokens comes at negligible additional latency. But once draft tokens are accepted,  multiple tokens are decoded with a single call to the target model, resulting in an overall latency reduction. Noticeably also, opposed to the compression techniques in \cref{sec:model_compression}, the output distribution provably remains the same \cite[Theorem 1]{chen:2023:speculative}. 

Beyond the verification also the draft token generation adds to the latency.
We classify SD methods broadly into two categories, based on whether or not they use a separate model for drafting. Seminal work \cite{Leviathan2023} uses a smaller model from the target model's family as draft model, e.g., T5-small as the draft model for T5-XXL, whereas \citet{Chen2023} train a separate draft model from scratch. 



Choosing an appropriate draft model for a target model can be tricky. In light of this, some SD methods take advantage of the target model itself. For example, self-speculative decoding \cite{self_spec} drafts tokens using the target model but skips some of its intermediate layers. Medusa \cite{medusa} trains multiple feed-forward heads on top of the last Transformer layer. The $i$-th head is responsible for predicting $(t+i)$-th token into the future. 
EAGLE~\cite{eagle} improves the heads by introducing auto-regression on features at the last Transformer layer. 
PaSS\cite{pass} appends $k$ special ``look-ahead" tokens to the prompt as input, and generates $k$ draft tokens in parallel using the target model itself. Lookahead Decoding~\cite{lookahead} applies Jacobi method \cite{Song2021,Santilli2023} that drafts multiple tokens in parallel. In some applications (e.g., Question Answering), one can draft tokens by matching their prefix in a document \cite{prefix_matching}, or a database~\cite{he2023rest}.

There are two orthogonal paths to further speed up speculative decoding. One is to draft multiple sequences for verification. The other is to improve the acceptance rate. We elaborate on them next.
\paragraph{Multiple Drafted Sequences}
In the vanilla case of a single drafted sequence, all drafted tokens after the first rejection position are wasted. In contrast, drafting multiple sequences increases the chance of having a longer accepted sub-sequence. The multiple sequences are often organized in a tree structure to share some prefixes. Correspondingly, the verification is made more efficient by introducing tree attention, with a specialized attention mask that reflects the token dependencies in the tree. This approach is first proposed in SpecInfer~\cite{specinfer}, adopted in several aforementioned papers (e.g., Medusa \cite{medusa}, EAGLE~\cite{eagle}, Lookahead Decoding~\cite{lookahead}), and further developed by \citet{sequoia}. Depending on the model architectures, speedups reported are often in $2-3\times$.

\paragraph{Aligning Draft to Target Model}
In~\cite{Leviathan2023}, the rejection rate is shown to be equal to the \emph{Total Variation} divergence (\emph{TV-div}) between target and draft models' token probabilities. 
This neat theoretical result has motivated Distillspec~\cite{distillspec} to knowledge distill from the target to draft model. With the better aligned draft model, $10-45$\% further speedups are reported. Regarding the objective function for distillation, it could be either conventional Kullback–Leibler divergence (\emph{KL-div}) or the more relevant \emph{TV-div}. Note that \emph{KL-div} can be considered as a surrogate for \emph{TV-div} due to Pinsker's inequality. Interestingly, \cite{distillspec} does not observe an obvious advantage of \emph{TV-div} against \emph{KL-div}.



\section{Conclusion}

This paper provides a comprehensive overview of efficient inference methods for LLMs, covering system optimization, structured Transformer architectures, model compression, and algorithmically faster decoding, especially in the context of AI accelerator. These techniques aim to facilitate effective computation, often considering input-output (IO) communication during attention score calculation, reducing extensive and repetitive self-attention mechanisms, minimizing memory idleness and compressing models themselves. 

Inference optimization is not only crucial for Transformer-based LLMs, but also other foundation models like 
Stable Diffusion~\citep{Rombach_2022_CVPR} or the Transformer alternative of State Space Models (SSMs)~\cite{Gu2022_SSM, gu2024mamba}. Several of the techniques presented in this paper have been successfully applied to these models too; e.g., in Stable Diffusion with FlashAttention~\cite{Chen2023_speed_diffusion}, quantization~\cite{shang2023posttraining,li2023qdiffusion,NEURIPS2023_2aab8a76, wang2023towards}, sparsity~\cite{li2023efficient}, and distillation~\cite{salimans2022progressive,meng2023distillation}, or in SSMs with Mixture of Experts~\cite{ anthony2024blackmamba, pioro2024moemamba}.  

Nevertheless, 
many of the
challenges remain largely unresolved, particularly when dealing with extremely long context lengths and sequences, necessitating tailored efforts depending on the types of devices used. We are confident that researchers and developers will continue to strive towards narrowing these gaps, thereby enhancing the accessibility of Generative AI systems.



\bibliographystyle{ACM-Reference-Format}
\balance
\bibliography{references}
\end{document}